\documentclass[conference]{IEEEtran}
\IEEEoverridecommandlockouts
\usepackage{cite}
\usepackage{amsmath,amssymb,amsfonts}
\usepackage{algorithmic}
\usepackage{graphicx}
\usepackage{textcomp}
\usepackage{xcolor}
\usepackage{amsmath}
\usepackage{color}
\def\BibTeX{{\rm B\kern-.05em{\sc i\kern-.025em b}\kern-.08em
    T\kern-.1667em\lower.7ex\hbox{E}\kern-.125emX}}
\begin{document}

\title{A Spatio-Temporal Attention-Based Method for Detecting Student Classroom Behaviors\\

\thanks{\textbf{\textcolor{red}{I finished this paper in the second half of 2021, and it was written in Chinese at that time.}} Now I have translated it into English and published it in arixv. Here is the Chinese manuscript of my thesis and the link to the relevant materials of my graduation thesis: https://share.weiyun.com/9qOxwpqz}
}

\author{\IEEEauthorblockN{1\textsuperscript{st} FAN YANG}
\IEEEauthorblockA{\textit{Jinan University} \\
Guangzhou, China \\
winstonyf@qq.com}

\and
\IEEEauthorblockN{2\textsuperscript{nd} Xiaofei Wang}
\IEEEauthorblockA{\textit{Chengdu University} \\
Chengdu, China \\
wangxiaofei@cdu.edu.cn}

\and
\IEEEauthorblockN{3\textsuperscript{rd} Baocai Zhong}
\IEEEauthorblockA{\textit{Chengdu Neusoft University} \\
Chengdu, China \\
zhongbaocai@nsu.edu.cn}

\and
\IEEEauthorblockN{3\textsuperscript{rd} Deyu Liu}
\IEEEauthorblockA{\textit{Chengdu Neusoft University} \\
Chengdu, China \\
LiuDeyu@nsu.edu.cn}

\and
\IEEEauthorblockN{4\textsuperscript{th} Jialing	Zhong}
\IEEEauthorblockA{\textit{Chengdu Neusoft University} \\
Chengdu, China \\
zhongjialing@nsu.edu.cn}

\and
\IEEEauthorblockN{5\textsuperscript{th} Tao Wang}
\IEEEauthorblockA{\textit{Chengdu Neusoft University} \\
Chengdu, China \\
wang-tao@nsu.edu.cn}

}
\maketitle

\begin{abstract}
Accurately detecting student behavior from classroom videos is beneficial for analyzing their classroom status and improving teaching efficiency. However, low accuracy in student classroom behavior detection is a prevalent issue. We propose a Spatio-Temporal Attention-Based Method for Detecting Student Classroom Behaviors (BDSTA) to address this issue. Firstly, the SlowFast network generates motion and environmental information feature maps from the video. Then, the spatio-temporal attention module is applied to the feature maps, including information aggregation, squeeze, and excitation processes. Subsequently, attention maps in the time, channel, and space dimensions are obtained, and multi-label behavior classification is performed based on these attention maps. To solve the long-tail data problem in student classroom behavior datasets, we use an improved focal loss function to assign more weight to the tail class data during training. Experimental results are conducted on a self-made student classroom behavior dataset named STSCB. Compared with the SlowFast model, the average accuracy of student behavior classification detection improves by 8.94\% using BDSTA.
\end{abstract}

\begin{IEEEkeywords}
 behavior detection, spatio-temporal attention module, long-tail data, focal loss function
\end{IEEEkeywords}

\section{Introduction}
Through the use of behavior detection technology \cite{b1}, accurate detection of student behavior in class videos can lead to the analysis of their classroom performance and learning behavior, which has significant implications for teachers, administrators, students, and their families. In traditional teaching, teachers find it challenging to focus on all students in the classroom and require feedback on their teaching methods by observing only a small number of students. School administrators can identify issues in their education and teaching methods only by directly observing classroom settings and analyzing student performance reports. Similarly, parents can only indirectly understand their child's learning behavior through feedback from teachers and their child. Accordingly, many researchers have adopted computer vision technology to automatically detect student behavior in class, such as Ngoc et al. \cite{b2}, who proposed a computer vision-based class behavior detection system that involves monitoring, behavior detection, and output feedback.

\begin{figure}
\centerline{
\includegraphics[width=0.5\textwidth]{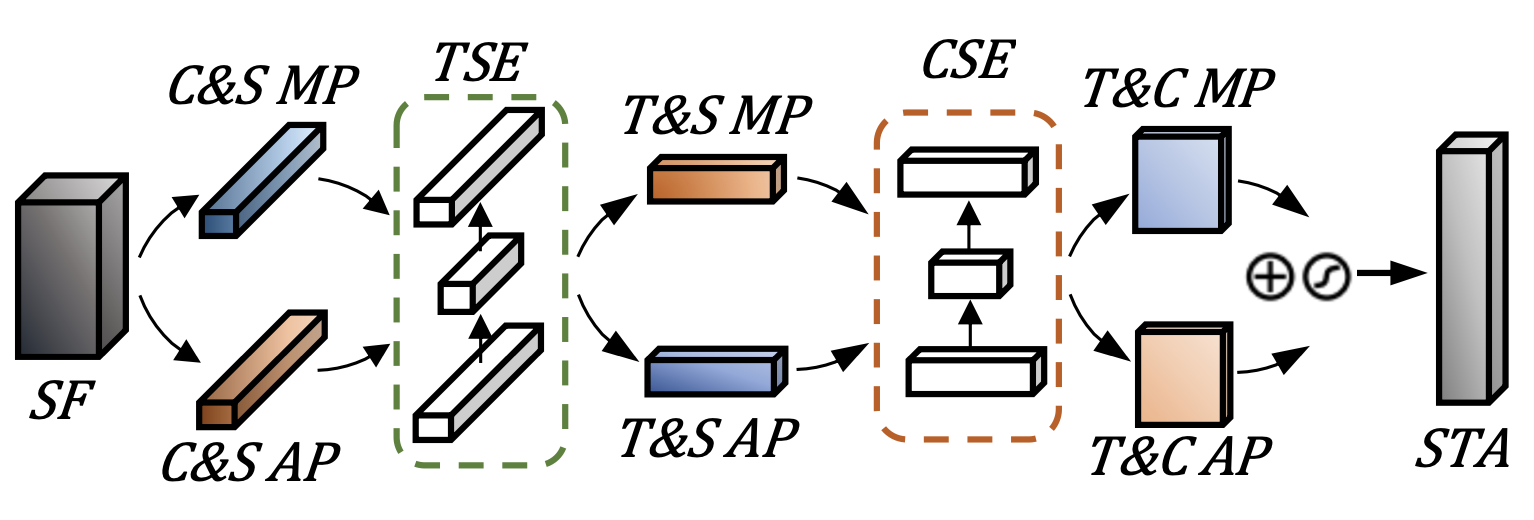}}
\caption{TCS3D Temporal, Channel, Spatial Convolution Attention Module.}
\label{netS}
\end{figure}

Currently, most class behavior detection is based on image detection\cite{Li_Student_behavior, BiTNet, CBPH-Net}, which has the advantages of small data annotation requirements and a small computation load. However, it is difficult to accurately judge continuous movements, such as identity association and behavior detection of students, as seen in class behavior detection based on Faster R-CNN \cite{b3}. With the significant progress made in action recognition \cite{b4}, temporal behavior detection \cite{b5}, and spatio-temporal behavior detection \cite{Slowfast,Videomae}, video-based class behavior detection has become the preferred method, such as class behavior recognition based on deep spatio-temporal residual convolutional networks \cite{b7}. This involves extracting spatio-temporal behavior features from student videos through a deep learning network to some extent, addressing the issue of detecting behavior when there are many students and significant occlusion in classroom settings.

However, using video-based student behavior datasets for class behavior detection requires the annotation of large amounts of sample data. The quantity of each behavior class will vary, causing the sample quantities for each behavior class to be unbalanced, where the majority of samples belong to the minority class (head class), and the majority class (tail class) possesses fewer samples. This data distribution is often referred to as a long-tail data distribution \cite{b8}.

Currently, video-based behavior datasets can be divided into temporal behavior datasets and spatio-temporal behavior datasets. Temporal behavior data only requires the behavior class and the time when it occurred in the video, such as UCF101 \cite{b9} and HMDB \cite{b10}. Spatio-temporal behavior data, on the other hand, requires the location of the related behavior target, such as AVA \cite{b11} and VATEX \cite{b12}. However, there is a lack of open temporal/spatio-temporal behavior datasets in the education field, which severely limits the application of video behavior detection in this area. Therefore, by observing student class behavior and conducting relevant literature research \cite{b13}, this paper constructs a spatio-temporal student class behavior dataset (STSCB).

There are three strategies to solve the long-tail data issue in the student class behavior dataset: resampling, weight assignment, and model fine-tuning. Resampling involves undersampling the head class and oversampling the tail class \cite{b14}, but it may lead to problems such as overfitting of the model or insufficient feature expression. Weight assignment involves setting the weight of each class in the loss function, assigning a larger weight to low prediction values and a smaller weight to high prediction values, thereby strengthening the loss function's impact on the tail class, such as weighted loss inter-class distance optimization based on sample numbers \cite{b15}, weighted based on effective sample numbers \cite{b16}, but weight assignment is sensitive to selecting hyperparameters and requires a significant number of experiments to determine optimal values. Model fine-tuning involves using a smaller learning rate, group aggregation, and other methods \cite{b17} to fine-tune the trained model, but this method has a high computation load and weak generalization ability. As BDSTA is based on the STSCB dataset for behavior detection, resampling methods are unsuitable due to multiple-label video attributes, and model fine-tuning methods are unsuitable for video detection as its computation load is higher than image detection. Therefore, this paper proposes an improved focal loss function by merging binary cross-entropy loss function and focal loss function, aiming to solve the low accuracy issues brought by long-tail data.

Extending the scale of the video detection network in time, channel, and space dimensions \cite{b4} can improve the model's detection accuracy to a certain extent. SlowFast \cite{Slowfast}, for example, uses different time, space, and channel extensions in the slow and fast paths to obtain environment and motion information. Tran et al. \cite{b18} tested different channel interrelationships to achieve good video detection accuracy at low computation. Wang et al. \cite{b19} established frame-to-frame matching between different layers of convolutional feature maps in the network to fuse appearance features and temporal information. There are currently many video detection network structures, many of which are extended from 2D image detection network structures into 3D spatio-temporal detection networks. For example, ResNet and Inception 2D models are directly extended into 3D models \cite{b20}. Furthermore, the extension of the 2D model to handle optical flow information in the video has been achieved \cite{b21}. Howard et al. \cite{b22} improved the computational efficiency of the model by increasing or decreasing the size of the input feature map. Mnasnet \cite{b23}, on the other hand, designed a spatial dimension network by incorporating a squeeze and excitation module \cite{b24} (SE), which obtains global information of the feature map through global average pooling and learns the correlation between channels through a fully connected layer. The convolutional block attention module (CBAM) \cite{b25} focuses on spatial information based on SE by using channel attention modules to enhance feature maps, followed by entering enhanced feature maps to space attention modules. CBAM enhances the model's ability to extract features with only a small number of parameters increased. This paper extends CBAM to the 3D network, and a 3D time, channel, and space convolutional block attention module (TCS3D) is proposed to improve the model's detection accuracy by extending the model in multiple dimensions. As shown in Figure 1, TCS3D is attached to SlowFast. SF represents the feature map of the video after being processed by the SlowFast network. C\&S stands for channel and space, T\&S represents time and space, and T\&C represents time and channel. MP and AP stand for max and average pooling, respectively. TSE represents time dimension squeeze and excitation, CSE represents the gain of the channel dimension after aggregation based on the global average pooling, and ST represents the spatio-temporal attention map.

The main contributions of this paper include: 

1) Construction of the spatio-temporal student class behavior dataset STSCB, which summarizes eight common behavior classes in the classroom: sitting, reading/writing, turning head/body, using the phone, leaning on the desk, raising-hand, standing/walking, and talking.

2) Proposing an improved focal loss function that solves the problem of low accuracy in long-tail data by fusing binary cross-entropy loss and focal loss to reduce the weight of the head class data in the loss function and increase the weight of the tail class in the loss function.

3) Proposing an improved attention module, TCS3D, which focuses on and excites temporal or channel information to obtain weight information of time and space and to concatenate spatial weight information. Through the extension of the network in multiple dimensions, it enhances the model's detection accuracy.

\begin{figure*}
\centerline{
\includegraphics[width=0.9\textwidth]{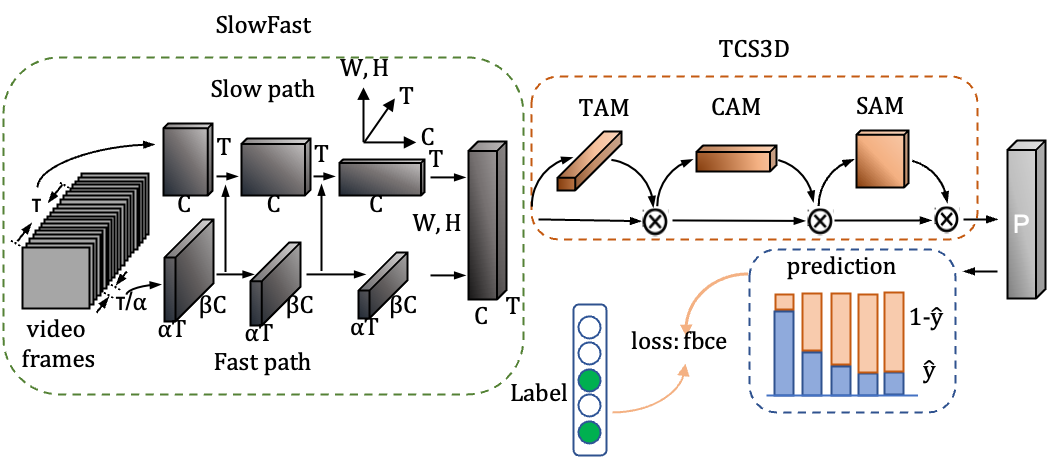}}
\caption{The network structure of the student classroom behavior detection based on spatio-temporal attention.}
\label{DBSTA}
\end{figure*}

\section{Dataset Construction}

The raw video materials for the dataset were sourced from 20 students from Sichuan Normal University who independently filmed classroom behaviors.  Each video contains 1 to 5 students, and scenes include classrooms, laboratories, and study rooms.  The videos have an average duration of about 90 seconds.  The videos were cropped using ffmpeg\\cite{FFmpeg} tool, with 4 frames cropped per second.

The STSCB data set consists of 8 behavior classes: sitting, reading/writing, turning head/body, using the phone, leaning on the desk, raising-hand, standing/walking, and talking.  STSCB is a spatio-temporal behavior detection data set, with behavior classes classified by the spatial position and behavior of the student.  Student spatial positions were detected using faster rcnn\cite{R-CNN}, while student behavior was labeled using via\cite{VIA}.

To address the problem of insufficient data samples, data augmentation was used to expand the data set.  Video frames were subject to salt and pepper noise, Gaussian noise, histogram equalization, Laplacian algorithm enhancement, and gamma transformation processing.  Each video was randomly processed using 3 of the abovementioned methods, resulting in 60 processed videos per video.

\begin{figure}
\centerline{
\includegraphics[width=0.46\textwidth]{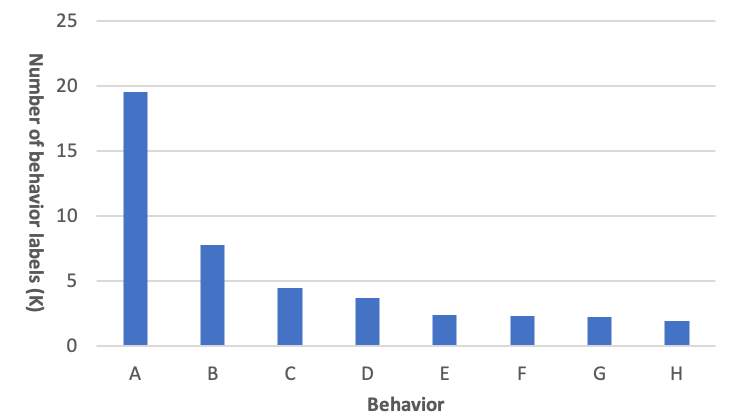}}
\caption{Number of labels for each class in the dataset.}
\label{Number_of_labels}
\end{figure}

STSCB contains a total of 24,628 video frames, with 44,670 behavior annotations.  The distribution of annotations for each class is shown in Fig.  \ref{Number_of_labels}

The classes are represented by letters A through H where sitting, using the phone, reading/writing, standing/walking, turning head/body, raising-hand, leaning on the desk, and talking are represented by A, B, C, D, E, F, G, and H, respectively.

\section{A Spatio-temporal Attention-based Detection Method}

SlowFast\cite{Slowfast} is a dual-stream network structure. One stream processes the semantic information of the video's spatial features at a low frame rate (Slow path, with frames input every $\tau$ frames, where $\tau =16$). The other stream processes the motion information at a high frequency (Fast path, with frames input every $\tau/\alpha$ frames, where $\alpha=8$). As shown on the left side of Fig. \ref{DBSTA}, the feature channels for the Slow path are represented by C, while the Fast path has $\beta C$ feature channels, where $\beta=1/8$, allowing the Fast path to run at a higher speed. In terms of time dimensions, the Slow path has a time dimension of T, while the Fast path has $\alpha$ T to better capture motion information. The paths share the same H and W dimensions to allow for lateral connections. SlowFast merges the features of the dual-stream branches via multiple lateral connections, and the merged information is then fed into a classifier for multi-label classification prediction.

BDSTA takes the feature map obtained after the final lateral connection of SlowFast as the input feature map $F\in R^{(C\times T\times H\times W)}$. The TCS3D attention module is then added after this feature map. Each module focuses on the information in the video feature from the perspectives of "when," "what," and "where." The formula for TCS3D is as follows:

\begin{equation}
    \centering
    \begin{aligned}
        F^\prime=M_t\ (F)\ \ \bigotimes\ F
    \end{aligned}
    \label{TCS3D1}
\end{equation}

\begin{equation}
    \centering
    \begin{aligned}
        F^{\prime\prime}=M_c\ (F^\prime\ )\ \ \bigotimes\ F^\prime
    \end{aligned}
    \label{TCS3D2}
\end{equation}

\begin{equation}
    \centering
    \begin{aligned}
        F^{\prime\prime\prime}=M_s\ (F^\prime\prime\ )\ \ \bigotimes F^{\prime\prime}
    \end{aligned}
    \label{TCS3D3}
\end{equation}

Where $M_t\in R^{(1\times T\times 1\times 1)}$ represents the 1D temporal attention map, $M_c\in R^{(C\times 1\times 1\times 1)}$ represents the 1D channel attention map, and $M_s\in R^{(1\times 1\times H\times W)}$ represents the 2D spatial attention map. The symbol $\bigotimes$ denotes element-wise multiplication. During the three multiplications, attention is broadcasted accordingly: the temporal attention values are broadcasted along the channel and spatial dimensions, and the channel attention values are broadcasted along the spatial dimension. $F^{\prime\prime\prime}$ represents the final output feature map.

The structure of TCS3D is shown in the right part of Fig. \ref{DBSTA}, consisting of the Time Attention Module (TAM), Channel Attention Module (CAM), and Space Attention Module (SAM) concatenated together. In TCS3D, channel/time and spatial information are focused by performing max pooling and average pooling operations on the channel/time and spatial dimensions, respectively. Then, the squeezed and excited operations are applied on the time/channel dimension to obtain the weight information for time/channel. Similarly, max pooling and average pooling operations are performed on the channel and time dimensions to focus on the channel and time information, respectively. Finally, the concatenated feature map is squeezed to obtain the weight information for space.

In the experiment, it was found that the STSCB dataset suffers from a severe data imbalance problem. An improved focal loss function, represented as FBce (Focal Bce Loss), was used in BDSTA to address this issue. As shown in Fig. \ref{DBSTA}, $\hat{y}$ represents the predicted value, and in the labels, solid circles represent the correct labels.

\subsection{TCS3D attention module}

After the lateral connections of spatial and motion information (generating a 5-dimensional feature map) in the SlowFast network by the BDSTA, the TCS3D attention module is incorporated.

In CBAM, the channel attention module and spatial attention module are used to process the feature maps of images. For feature maps of videos, there are 5 dimensions: N × C × T × W × H, where N represents the batch size, C represents the number of channels, T represents the time dimension, W represents the width, and H represents the height. Compared to feature maps of images, feature maps of videos have an additional time dimension, T.

In the experiments, two approaches were attempted to handle the additional time dimension: the first approach extended the time dimension based on CBAM, replacing the 2D convolutions in CBAM with 3D convolutions; the second approach applied PCA dimensionality reduction to the feature maps of videos, reducing the 5-dimensional feature maps to 4 dimensions. Experimental analysis showed that the second approach resulted in the loss of some feature information, leading to degraded detection performance.

\textbf{3D Temporal Attention Module}

\begin{figure}
\centerline{
\includegraphics[width=0.46\textwidth]{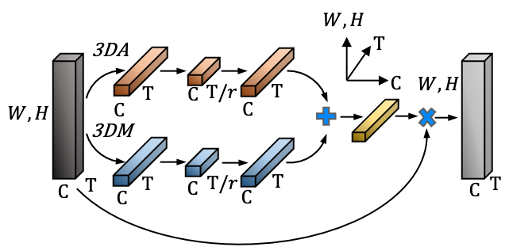}}
\caption{3D Temporal Attention Module.}
\label{TAM}
\end{figure}

The TAM utilizes the temporal information of features to generate a temporal attention map. The temporal attention focuses on the "when" aspect of the input video that is important. To effectively compute the temporal attention, the TAM incorporates 3D global average pooling and 3D global maximum pooling to capture channel and spatial information from the input feature map and then squeeze and excite the temporal dimension. The detailed process is shown in Fig. \ref{TAM}.

\begin{table}
    \scriptsize
    \centering \renewcommand\arraystretch{1.25} 
    \caption{Network architecture of 3D temporal attention module.}
    \label{Network_TAM}
    \begin{tabular}{ccc}
        \hline
        stage & Output network architecture & Description \\ \hline
        Input feature map & $C \times T \times W \times H$ & - \\
        3DA/3DM & $1 \times T \times 1 \times 1$ & Feature aggregation \\
        3D Conv 1 & $1 \times \frac{T}{r} \times 1 \times 1 \quad r=8$ & Squeeze \\
        3D Conv 2 & $1 \times T \times 1 \times 1$ & Excitation \\
        Merge & $1 \times T \times 1 \times 1$ & Element-wise sum \\
        Multiply & $C \times T \times W \times H$ & Feature map multiplication \\
        \hline
    \end{tabular}
\end{table}

In Fig. \ref{TAM}, 3DA represents 3D global average pooling, and 3DM represents 3D global maximum pooling. The plus sign denotes element-wise addition between the two feature maps, and the multiplication sign denotes element-wise multiplication. The network structure of the Temporal Attention Module (TAM) is shown in Table \ref{Network_TAM}. First, TAM uses 3D global average pooling and 3D global maximum pooling to aggregate channel and spatial information, generating two different 3D feature maps: $F_{avg}^t\in\mathbb{R}^{1\times T\times1\times1}$ and $F_{max}^t\in\mathbb{R}^{1\times T\times1\times1}$. Then, TAM feeds these two 3D feature maps into a shared network to produce the temporal attention map $M_t\in\mathbb{R}^{1\times T\times1\times1}$. The shared layer consists of a Multi-Layer Perceptron (MLP) and a hidden layer, where the excitation size of the hidden layer is set to $\mathbb{R}^{1\times T/r\times1\times1}$, where r represents the squeezed rate, and r=8 is set in the experiments. After the squeeze and excitation in the shared layer, TAM performs element-wise addition between the two feature maps. Finally, TAM multiplies the element-wise sum with the input feature map to obtain a temporal attention map with the same size as the input feature map.

The formula for TAM is as follows:

\begin{equation}
    \centering
    \begin{aligned}
        M_t(F) = \sigma \left( MLP \left( 3DA(F) \right) \right) + MLP \left( 3DM(F) \right) \\
        M_t(F) = \sigma \left( W_1 \left( W_0 \left( F_{avg}^t \right) \right) + W_1 \left( W_0 \left( F_{max}^t \right) \right) \right) 
    \end{aligned}
    \label{TAM_formula}
\end{equation}

Where $\sigma$ represents the sigmoid function, F represents the input feature map, and $W_0 \in \mathbb{R}^{C/r \times C}, W_1 \in \mathbb{R}^{C \times C/r}$.

\textbf{3D Channel Attention Module}

\begin{figure}
\centerline{
\includegraphics[width=0.46\textwidth]{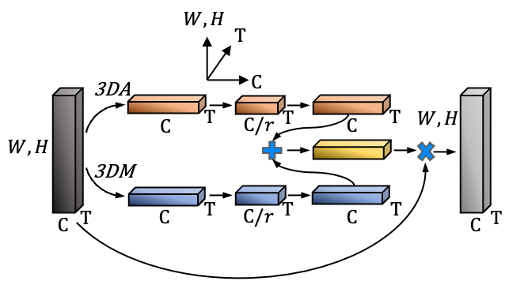}}
\caption{3D Channel Attention Module.}
\label{CAM}
\end{figure}

The CAM generates a 3D channel attention map by utilizing the channel relationships of features. The channel attention focuses on "what" is meaningful in the input video. In order to compute the channel attention effectively, TAM adopts 3D global average pooling and 3D global maximum pooling to capture temporal and spatial information, followed by channel-wise squeeze and excitation. The detailed process is shown in Fig. \ref{CAM}.

\begin{table}
    \scriptsize
    \centering \renewcommand\arraystretch{1.25} 
    \caption{Network architecture of 3D Channel Attention Module.}
    \label{Network_CAM}
    \begin{tabular}{ccc}
        \hline
        stage & Output network architecture & Description \\ \hline
        Input feature map & $C \times T \times W \times H$ & - \\
        3DA/3DM & $C \times 1 \times 1 \times 1$ & Feature aggregation \\
        3D Conv 1 & $\frac{C}{r} \times  1 \times 1 \times 1 \quad r=16$ & Squeeze \\
        3D Conv 2 & $C \times 1 \times 1 \times 1$ & Excitation \\
        Merge & $C \times 1 \times 1 \times 1$ & Element-wise sum \\
        Multiply & $C \times T \times W \times H$ & Feature map multiplication \\
        \hline
    \end{tabular}
\end{table}

The network structure of the Channel Attention Module (CAM) is shown in Table \ref{Network_CAM}. First, CAM utilizes 3D global average pooling and 3D global maximum pooling to aggregate temporal and spatial information, generating two different 3D feature maps: $F_{avg}^c\in\mathbb{R}^{C\times1\times1\times1} $ and $F_{max}^c\in\mathbb{R}^{C\times1\times1\times1}$. Then, CAM sends these two 3D feature maps into a shared network to generate a channel attention map $M_c\in\mathbb{R}^{C\times1\times1\times1}$. The size of the excitation in the hidden layer of the shared layer is set to $\mathbb{R}^{C/r\times1\times1\times1}$, where r=16 is set in the experiments. After passing through the shared layer, the two feature maps in TAM are element-wise summed. Finally, CAM multiplies the element-wise summed feature map with the input feature map to obtain a time attention map of the same size as the input feature map.

The formula for CAM is as follows:

\begin{equation}
    \centering
    \begin{aligned}
    M_c\left(F\right)=\sigma\left(MLP\left(3DA\left(F\right)\right)+MLP\left(3DM\left(F\right)\right)\right) \\
        M_c\left(F\right)=\sigma\left(W_1\left(W_0\left(F_{avg}^c\right)\right)+W_1\left(W_0\left(F_{max}^c\right)\right)\right)
    \end{aligned}
    \label{CAM_formula}
\end{equation}

\textbf{3D Spatial Attention Module}

The Spatial Attention Module (SAM) utilizes the spatial relationships of features to generate a spatial attention map. This spatial attention map focuses on "where" in the input video is meaningful. To effectively compute spatial attention, SAM employs 3D global average pooling and 3D global max pooling, taking into account both temporal and spatial information. The two generated feature maps are then concatenated along the channel dimension and passed through a $3\times3\times3$ 3D convolution operation to reduce the feature map to a single channel dimension. Finally, a spatial attention map $ M_s\in\mathbb{R}^{1\times1\times H\times W}$ is generated. The detailed process is illustrated in Figure 6.

\begin{figure}
\centerline{
\includegraphics[width=0.46\textwidth]{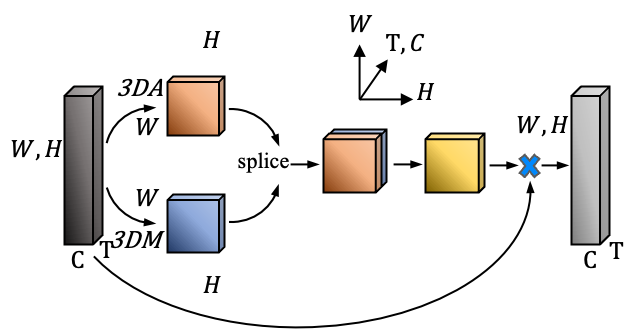}}
\caption{3D Spatial Attention Module.}
\label{SAM}
\end{figure}

\begin{table}
    \scriptsize
    \centering \renewcommand\arraystretch{1.25} 
    \caption{Network architecture of 3D Spatial Attention Module.}
    \label{Network_SAM}
    \begin{tabular}{ccc}
        \hline
        stage & Output network architecture & Description \\ \hline
        Input feature map & $C \times T \times W \times H$ & - \\
        3DA/3DM & $ 1 \times 1 \times W \times H$ & Feature aggregation \\
        splice & $ 2 \times 1 \times W \times H$ & Channel-based splicing \\
        3D Conv & $ 1 \times  1 \times W \times H$ & Squeeze \\
        Merge & $1 \times 1 \times W \times H$ & Element-wise sum \\
        Multiply & $C \times T \times W \times H$ & Feature map multiplication \\
        \hline
    \end{tabular}
\end{table}

The network structure of the Spatial Attention Module (SAM) is shown in Table \ref{Network_SAM}. Firstly, SAM utilizes two pooling operations to aggregate temporal and channel information, generating two different 3D feature maps: $F_{avg}^s\in\mathbb{R}^{1\times1\times H\times W}$ and $F_{max}^s\in\mathbb{R}^{1\times1\times H\times W}$. Then, the two feature maps are concatenated and passed through a 3D convolution operation. Finally, the resulting feature map is multiplied element-wise with the original input feature map, producing a spatial attention map of the same size as the input feature map.

The formula for SAM is as follows:

\begin{equation}
    \centering
    \begin{aligned}
    M_s\left(F\right)=\sigma\left(f^{3\times3\times3}\left(\left[3DA\left(F\right);3DM\left(F\right)\right]\right)\right) \\
        M_s\left(F\right)=\sigma\left(f^{3\times3\times3}\left[F_{avg}^s;F_{max}^s\right]\right)
    \end{aligned}
    \label{SAM_formula}
\end{equation}

Where $F^{3\times3\times3}$ represents a 3D convolution operation, and $3\times3\times3$ represents the size of the convolution kernel.

\subsection{Improved Focal Loss}
In the STSCB dataset, the distribution of behavior class labels is shown in Fig. \ref{Number_of_labels}. The two head classes, "sitting" and "using the phone," account for 61$\%$ of the total annotations, while the remaining 6 tail classes account for 39$\%$ of the total annotations. During the model's training process, the model tends to favor the head classes, increasing the error rate for the tail classes.

In spatio-temporal behavior detection and classification, the commonly used loss function for multi-label classification is binary cross-entropy loss (BCE Loss). However, BCE Loss does not consider the differences in contribution between head and tail classes in the long-tailed data, resulting in low model accuracy during training. Therefore, BDSTA adopts an improved focal loss function by combining focal loss and BCE Loss, proposing the FBce (Focal BCE Loss) to reduce the weight of the loss function for head classes and increase the weight of the loss function for tail classes in multi-label data. The formula for FBce is as follows:

\begin{equation}
    \centering
    \begin{aligned}
        &A=-\alpha L\left(1-\hat{y}\right)^\gamma L\left(\hat{y}\right)y 
        \\
        &B=\left(1-\alpha\right){\hat{y}}^\gamma L\left(1-\hat{y}\right)\left(1-y\right)
        \\
        &F\_BL=A-B
    \end{aligned}
    \label{FBce_formula}
\end{equation}

Where L represents the natural logarithm function, $\alpha$ represents the weighting factor, $\alpha\in\left[0,1\right]$, where the positive class is $\alpha$, and the negative class is $1-\alpha$. $y$ represents the correct label. $\hat{y}$ represents the probability of y=1, $\hat{y}\in\left(0,1\right)$. $\gamma$ is the focusing parameter, $\gamma\in\left[0,5\right]$, where$ \left(1-\hat{y}\right)^\gamma$ is the modulation factor.

When $\hat{y}\rightarrow0$, the modulation factor $\left(1-\hat{y}\right)^\gamma $ approaches 1, so the weight of correctly classified samples increases. When $\hat{y}\rightarrow1$, the modulation factor $\left(1-\hat{y}\right)^\gamma $ approaches 0, and the weight of correctly classified samples decreases.

By adjusting the focusing parameter $\gamma$, we can reduce the weight of easily classified samples. When $ \gamma=0$, FBce is equivalent to BceLoss. As $\gamma $increases, the influence of the modulation factor $\left(1-\hat{y}\right)^\gamma $also increases. In the experiments of BDSTA, $\gamma=5 $achieves the best performance.

\begin{figure*}
\centerline{
\includegraphics[width=0.755\textwidth]{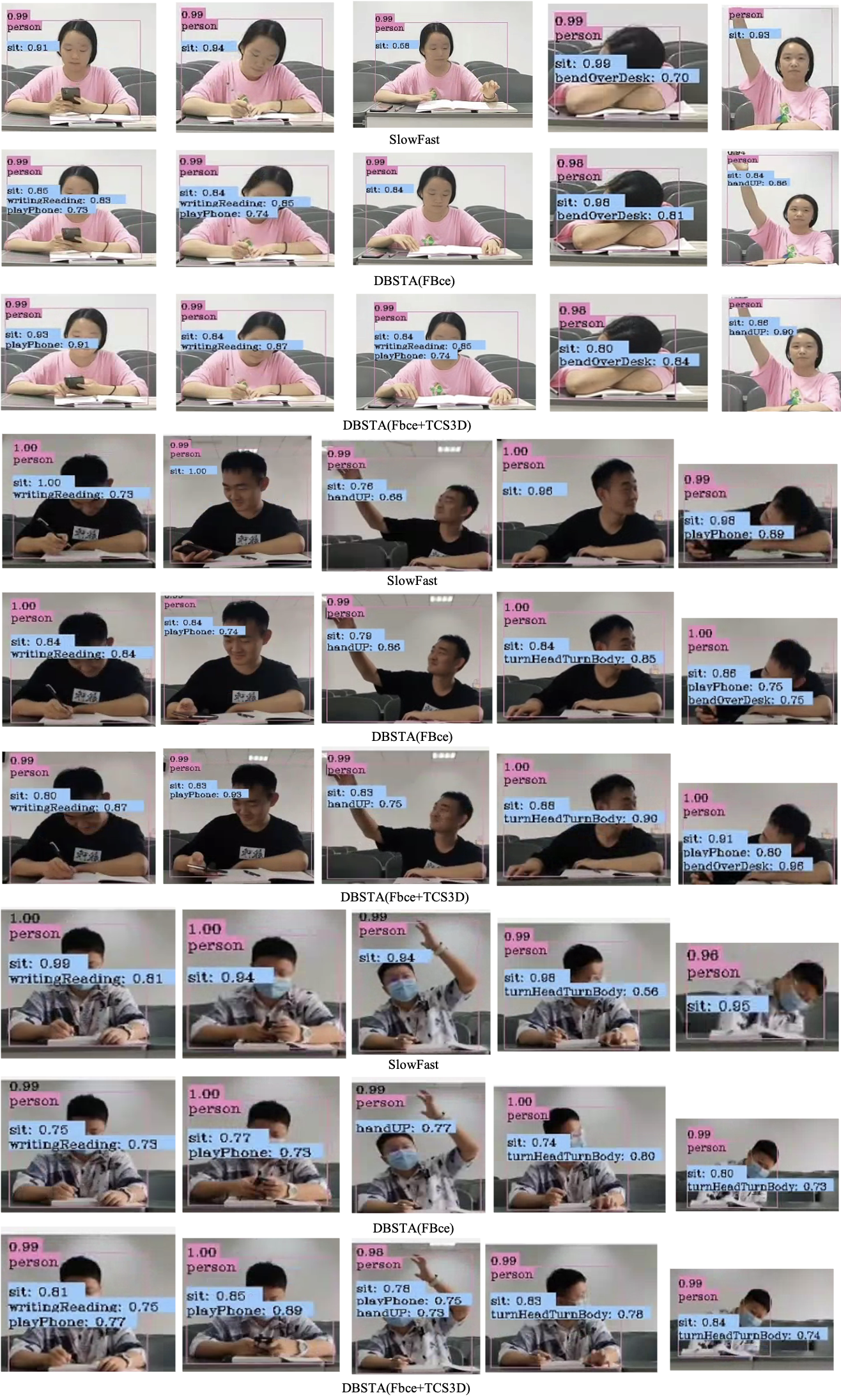}}
\caption{Comparison of detection results of three models.}
\label{DBSTA_sample}
\end{figure*}

\section{Experimental Results and Analysis}

\subsection{Experimental Environment and Dataset}
The GPU used for model training is NVIDIA GeForce RTX 2080 Ti with 11GB of memory. The operating system is Ubuntu 20.04.2. The programming language is Python 3.6.13. The version of PyTorch used is 1.8.1. The CUDA version is 10.1. The dataset used in the experiment is STSCB, with a ratio of 4:1:1 for the training set, test set, and validation set.

\subsection{Experimental Content}
To verify the effectiveness of the proposed BDSTA, the experiments are conducted from the following three aspects:
1) Comparing the detection results of the model before and after the improvement of the loss function.
2) Comparing the detection results of the model before and after incorporating TCS3D.
3) Comparing the detection results of the model before and after incorporating the improved loss function and TCS3D.

\subsection{Model Training}
The model training consists of three parts. The first part involves incorporating the improved focal loss function into BDSTA. The second part involves incorporating TCS3D into BDSTA. The third part involves incorporating the improved focal loss function and TCS3D into BDSTA. To improve training speed and achieve better convergence, all three models use a pre-trained SlowFast model.

The SGD optimizer is used for model training with a momentum of 0.9, weight decay of 0.00001, initial learning rate of 0.075, and 40 epochs.

\subsection{Evaluation Metrics}
This paper uses mean average precision (mAP) with an IOU threshold of 0.5 to objectively analyze the experimental results. The formula is as follows:

\begin{equation}
    \centering
    \begin{aligned}
        Recall=\frac{TP}{TP+FN}
    \end{aligned}
    \label{Recall}
\end{equation}

\begin{equation}
    \centering
    \begin{aligned}
        Precision=\frac{TP}{TP+FP}
    \end{aligned}
    \label{Precision}
\end{equation}

\begin{equation}
    \centering
    \begin{aligned}
        mAP=\int_{0}^{1}{P(R)dR}
    \end{aligned}
    \label{mAP}
\end{equation}

Equation \ref{Recall}: Recall is the ratio of true positive (TP) to the sum of TP and false negative (FN), denoted as R.

Equation \ref{Precision}: Precision is the ratio of TP to the sum of TP and false positive (FP), denoted as P.

TP (True Positive) represents the number of positive samples correctly classified as positive.

FN (False Negative) represents the number of positive samples incorrectly classified as negative.

FP (False Positive) represents the number of negative samples incorrectly classified as positive.

TP + FN represents the total number of positive samples.

TP + FP represents the total number of samples classified as positive.

TP and FP are determined based on the IOU (Intersection Over Union) threshold. The IOU calculation formula is as follows:

\begin{equation}
    \centering
    \begin{aligned}
        IOU(A,B)=\left|\frac{A\cap B}{A\cup B}\right|
    \end{aligned}
    \label{IOU}
\end{equation}

In the equation, A represents the ground truth box, and B represents the box predicted based on anchors and detection models. 

To further analyze the role of the improved loss function and TCS3D in the network, this paper adopts false detection rate and missing detection rate as evaluation metrics, with the following formulas:

\begin{equation}
    \centering
    \begin{aligned}
        FR=\frac{1}{N}\sum_{i=1}^{N}\frac{fp}{M}
    \end{aligned}
    \label{FR}
\end{equation}

\begin{equation}
    \centering
    \begin{aligned}
        MR=\frac{1}{N}\sum_{i=1}^{N}\frac{fn}{M}
    \end{aligned}
    \label{MR}
\end{equation}

In the equation, FR represents the false detection rate, MR represents the missing detection rate, fp represents the number of false positives where negative samples are incorrectly classified as positive in a video frame, and fn represents the number of false negatives where positive samples are incorrectly classified as negative in a video frame. M represents the total number of behavior classes; in the experiment, M is set to 8, and N represents the number of video frames input into the model.

\subsection{Experimental Results and Analysis}

\textbf{Experimental Results and Analysis of FBce Loss Function}

To verify the effectiveness of the improved focal loss function FBce in addressing long-tailed data, the value of the focus parameter gamma in FBce is set to 2, where gamma is related to the imbalance level of the dataset. The experiment found that the mAP increased from 73.61$\%$ to 80.0$\%$ and the missing detection rate decreased from 12.72$\%$ to 5.16$\%$ with the implementation of FBce.

\begin{figure}
\centerline{
\includegraphics[width=0.46\textwidth]{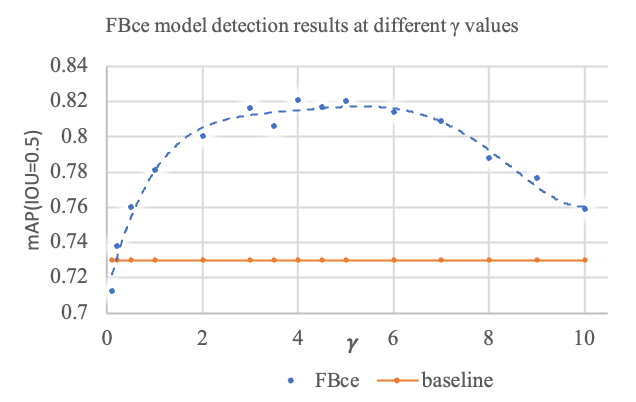}}
\caption{The effect of FBce taking different $\gamma$ on the model.}
\label{fbce_gamma}
\end{figure}

To analyze the impact of the focus parameter gamma in FBce on the model's detection results, different values of gamma are set from 0.1 to 10 in the experiment. The experimental results are shown in Fig. \ref{fbce_gamma}, where the solid line represents the baseline, which is the detection result of the BDSTA network using the Bce loss function with an mAP of 73.61$\%$. The dots represent the detection results of the BDSTA network using FBce, and the dashed line is a sixth-degree polynomial fitting curve to the dots. The horizontal axis represents the different values of the focus parameter gamma ranging from 0.1 to 10. It can be observed that when gamma=5, the model's detection results reach the optimal result of 82.04$\%$; for gamma values in the range of [0,2], the mAP improvement rate is the highest, and when $\gamma>6$, the mAP of the detection results starts to decline.

\begin{figure}
\centerline{
\includegraphics[width=0.46\textwidth]{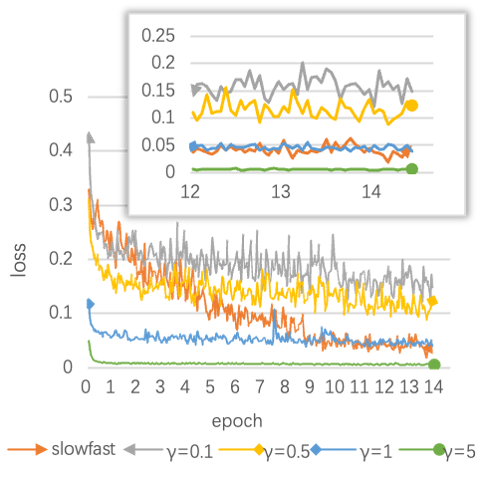}}
\caption{FBce takes the contrast curve of the model loss function under different $\gamma$.}
\label{Fbce_loss}
\end{figure}

To analyze the effect of the focal parameter $\gamma$ in FBce on the model training process, the experiment recorded the changes in the loss function in the training process under different $\gamma$ values. The experimental results are shown in Fig. \ref{Fbce_loss}, where the curve pointing to the right represents the loss function of the SlowFast training process, which serves as the reference line. The curve with the left-pointing square represents the loss function of the BDSTA training process using FBce with $\gamma=0.1$; the curve with the right-pointing square represents the loss function of the BDSTA training process using FBce with $\gamma=0.5$; the curve with the left-pointing circle represents the loss function of the BDSTA training process using FBce with $\gamma=1$; the curve with the right-pointing circle represents the loss function of the BDSTA training process using FBce with $\gamma=5$. It can be observed that during the training process, as $\gamma $ gradually increases in the BDSTA with FBce, $\gamma\in[0,5]$, the convergence of the loss function accelerates, and the loss value becomes lower and more stable.

\begin{figure}
\centerline{
\includegraphics[width=0.46\textwidth]{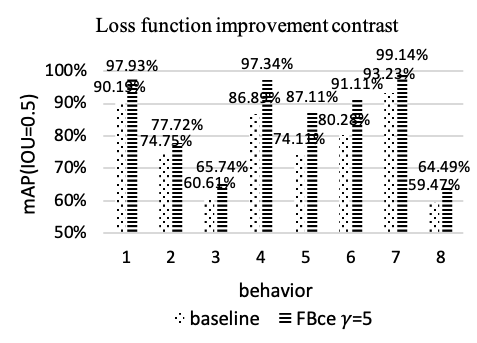}}
\caption{mAP value for model classes when $\gamma=5$ in FBce}
\label{FBce_gamma5}
\end{figure}

In order to further analyze the improvement of the detection results of different behaviors in the STSCB dataset using FBce, the experiment set the focal parameter $\gamma=5$ in FBce and compared the detection results of each behavior class between the original model and the improved model. The experimental results are shown in Fig. \ref{FBce_gamma5}, where the behavior classes "sitting", "reading/writing," "turning head/body," "using the phone," "leaning on the desk," "raising-hand," "standing/walking," "talking" are represented by numbers 1 to 8. The bar chart on the right represents the detection results of the 8 behavior classes in BDSTA using FBce. The bar chart on the left represents the detection results of the 8 behavior classes in SlowFast. It can be seen that the improved network has good improvement in all classes, especially in detecting the behavior of leaning on desk, with an improvement of about 13$\%$.

When $\gamma=5$, comparing the miss rate and false alarm rate with adding FBce to BDSTA are shown in Table \ref{FBce_gamma5_table}.

\begin{table}
    \scriptsize
    \centering \renewcommand\arraystretch{1.25} 
    \caption{Network architecture of 3D Channel Attention Module.}
    \label{FBce_gamma5_table}
    \begin{tabular}{ccc}
        \hline
         & Missing rate($\%$)  & False detection rate($\%$) \\ \hline
        SlowFast feature map & 12.72 & - \\
        FBce $\gamma=5$ & 5.16 & 24.11 \\
        TCS3D + FBce $\gamma=5$ & 4.66 & 19.11 \\
        \hline
    \end{tabular}
\end{table}

From Table \ref{FBce_gamma5_table}, it can be seen that the SlowFast model has a high miss rate. The reason is that the SlowFast model has good detection performance on the head class but serious misses on the tail class. When FBce is added, the weight of the loss function is biased towards the tail class, reducing the miss rate but increasing the false alarm rate for the tail class.

\textbf{Experimental Results and Analysis of TCS3D Attention Module}

To verify the effect of the spatio-temporal attention module TCS3D on student detection in the classroom, the experiment added TAM, CAM, SAM, and TCS3D to SlowFast separately. The experimental results are shown in Table \ref{4model_map}, where TAM, CAM, SAM, and TCS3D are represented by T, C, S, and TCS respectively. The detection results with the four attention modules added to BDSTA are shown in the rows without FBce. It can be found that adding TCS3D achieves the highest mAP, with an improvement of 1.58$\%$ compared to the baseline detection result of 73.61$\%$, followed by adding TAM with a 1.58$\%$ improvement.
		
\begin{table}
    \scriptsize
    \centering \renewcommand\arraystretch{1.25} 
    \caption{4 Types of attention modules on Model Checking Results (mAP).}
    \label{4model_map}
    \begin{tabular}{ccccc}
        \hline
        & T($\%$) & C($\%$) & S($\%$) & TCS($\%$) \\ \hline
        None FBce & 75.53 & 74.86 & 75.18 & 76.12 \\
        FBce $\gamma=1$ & 80.78 & 81.70 & 79.86 & 84.04 \\
        FBce $\gamma=5$ & 81.48 & 82.24 & 82.30 & 82.55 \\
        \hline
    \end{tabular}
\end{table}

\begin{figure}
\centerline{
\includegraphics[width=0.46\textwidth]{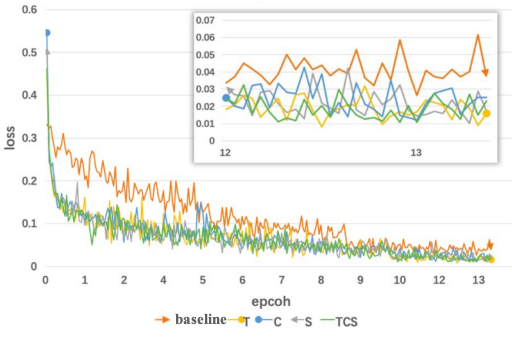}}
\caption{Comparison curve of model loss function under 4 attention modules}
\label{4modelLoss}
\end{figure}

In order to analyze the impact of the four attention modules on the training process of the model, the experiment recorded the changes in the loss function during the training process when the TCS3D attention module was added to BDSTA. The experimental results are shown in Fig. \ref{4modelLoss}, where the curve pointing to the right "slowfast" represents the loss function of the training process of SlowFast, which serves as the reference line; the curve with the right-pointing circle represents the loss function of the training process of BDSTA with TAM added; the curve with the left-pointing circle represents the loss function of the training process of BDSTA with TCS3D added; the curve with the left-pointing arrow represents the loss function of the training process of BDSTA with SAM added; the curve without any symbol represents the loss function of the training process of BDSTA with TCS3D added. From Fig. \ref{4modelLoss}, it can be seen that during the training process, adding TAM, CAM, SAM, and TCS3D can all accelerate the convergence speed to a certain extent.

From the above experiments, it can be seen that directly adding TCS3D to SlowFast does not significantly improve the detection rate of the model, which is due to the fact that TCS3D does not perform well under imbalanced data.

\textbf{Experimental Results and Analysis of FBce and TCS3D}

To further analyze the impact of adding FBce and TCS3D to BDSTA on model detection, it was found through the previous two experiments that adding TCS3D and setting $\gamma \in (0,5)$ yielded the best results. In the experiments, TAM, CAM, SAM, and TCS3D were added to BDSTA, and the value of the focus parameter $\gamma$ in FBce was set to 1 and 5. The experimental results are shown in Table \ref{4model_map}. It can be observed that when $\gamma=1$, the four attention modules showed more significant improvement, with an increase of 10.43$\%$ in mAP. When $\gamma=5$, the mAP increased by 8.94$\%$. However, compared to BDSTA with only FBce (with $\gamma=5$), the improvement was relatively small.

Comparing the miss rate and false alarm rate, it was found that by comparing BDSTA with only FBce (with $\gamma=5$) to BDSTA with FBce (with $\gamma=5$) and TCS3D, the false alarm rate decreased from 24.11$\%$ to 19.11$\%$ and the miss rate decreased from 5.16$\%$ to 4.66$\%$. The experiments indicated that TCS3D needs to be added to BDSTA together with FBce to achieve a significant improvement in model detection.

Fig. \ref{DBSTA_sample} shows the detection results of three different models on a subset of the STSCB dataset. The first row represents the detection results on SlowFast, the second row represents the detection results on BDSTA with FBce, and the third row represents the detection results on BDSTA with FBce and TCS3D. Each column represents a different video frame (different behaviors). It can be observed that SlowFast had more missed detections and relatively lower prediction accuracy. BDSTA (FBce) had significantly improved prediction accuracy compared to SlowFast and reduced the number of missed detections, but had relatively low prediction accuracy for some tail classes and a few false detections. BDSTA (FBce+TCS3D) further improved prediction accuracy compared to BDSTA (FBce), reduced the number of missed detections and false alarms, and further improved prediction accuracy for tail classes. By observing a large number of video detection results, it was found that using FBce in BDSTA can improve the detection of tail classes, but it also increases the false alarm rate for tail classes. However, using FBce with the addition of TCS3D in BDSTA can effectively reduce the false alarm rate for tail classes.

\section{Summary}
In this paper, we propose a student classroom behavior classification model based on spatio-temporal attention. This model learns the temporal, spatial, and channel features in the video to achieve multi-label behavior detection for multiple students in classroom scenarios. Additionally, the model utilizes an improved focal loss function to address the issue of long-tailed data in classroom scenarios, which improves the detection accuracy of the model while reducing both the miss rate and false alarm rate.

\vspace{12pt}


\begin{thebibliography}{00}
\bibitem{b1} GAO Chenqiang,CHEN Xu.Deep learning based action detection: a survey[J].Journal of Chongqing University of Posts and Telecommunications(Natural Science Edition),2020,32(06):991-1002.
\bibitem{b2} Ngoc Anh B, Tung Son N, Truong Lam P, et al. A computer-vision based application for student behavior monitoring in classroom[J]. Applied Sciences, 2019, 9(22): 4729.

\bibitem{Li_Student_behavior}Li Y, Qi X, Saudagar A K J, et al. Student behavior recognition for interaction detection in the classroom environment[J]. Image and Vision Computing, 2023: 104726.

\bibitem{BiTNet}Zhao J, Zhu H, Niu L. BiTNet: A lightweight object detection network for real-time classroom behavior recognition with transformer and bi-directional pyramid network[J]. Journal of King Saud University-Computer and Information Sciences, 2023, 35(8): 101670.

\bibitem{CBPH-Net}Zhao J, Zhu H. CBPH-Net: A Small Object Detector for Behavior Recognition in Classroom Scenarios[J]. IEEE Transactions on Instrumentation and Measurement, 2023.

\bibitem{b3} ZHOU Ye. Research on Classroom Behaviors Detection of Primary School Students Based on Faster R-CNN [D]. Sichuan Normal University,2021. DOI: 10.27347/d.cnki.gssdu.2021.000962
\bibitem{b4} Feichtenhofer C. X3d: Expanding architectures for efficient video recognition[C]//Proceedings of the IEEE/CVF Conference on Computer Vision and Pattern Recognition. 2020: 203-213.
\bibitem{b5} Lin T, Zhao X, Su H, et al. Bsn: Boundary sensitive network for temporal action proposal generation[C]//Proceedings of the European Conference on Computer Vision (ECCV). 2018: 3-19.
\bibitem{Slowfast} Feichtenhofer C, Fan H, Malik J, et al. Slowfast networks for video recognition[C]//Proceedings of the IEEE/CVF international conference on computer vision. 2019: 6202-6211.

\bibitem{Videomae}Tong Z, Song Y, Wang J, et al. Videomae: Masked autoencoders are data-efficient learners for self-supervised video pre-training[J]. Advances in neural information processing systems, 2022, 35: 10078-10093.

\bibitem{b7} HUANG Yongkang,LIANG Meiyu,WANG Xiaoxiao,CHEN Zheng,CAO Xiaowen.Multi-person classroom action recognition in classroom teaching videos based on deep spatiotemporal residual convolution neural network[J].Journal of Computer Applications,2022,42(03):736-742.
\bibitem{b8} Ouyang W, Wang X, Zhang C, et al. Factors in finetuning deep model for object detection with long-tail distribution[C]//Proceedings of the IEEE conference on computer vision and pattern recognition. 2016: 864-873.
\bibitem{b9} Soomro K, Zamir A R, Shah M. UCF101: A dataset of 101 human actions classes from videos in the wild[J]. arXiv preprint arXiv:1212.0402, 2012.
\bibitem{b10} Kuehne H, Jhuang H, Garrote E, et al. HMDB: a large video database for human motion recognition[C]//2011 International conference on computer vision. IEEE, 2011: 2556-2563.
\bibitem{b11} Gu C, Sun C, Ross D A, et al. Ava: A video dataset of spatio-temporally localized atomic visual actions[C]//Proceedings of the IEEE Conference on Computer Vision and Pattern Recognition. 2018: 6047-6056.
\bibitem{b12} Wang X ,  Wu J ,  Chen J , et al. VATEX: A Large-Scale, High-Quality Multilingual Dataset for Video-and-Language Research[J]. IEEE, 2019.
\bibitem{b13} Sun B ,  Y  Wu,  Zhao K , et al. Student Class Behavior Dataset: a video dataset for recognizing, detecting, and captioning students' behaviors in classroom scenes[J]. Neural Computing and Applications, 2021:1-20.
\bibitem{b14} Pouyanfar S, Tao Y, Mohan A, et al. Dynamic sampling in convolutional neural networks for imbalanced data classification[C]//2018 IEEE conference on multimedia information processing and retrieval (MIPR). IEEE, 2018: 112-117.
\bibitem{b15} Cao K, Wei C, Gaidon A, et al. Learning imbalanced datasets with label-distribution-aware margin loss[J]. arXiv preprint arXiv:1906.07413, 2019.
\bibitem{b16} Cui Y ,  Jia M ,  Lin T Y , et al. Class-Balanced Loss Based on Effective Number of Samples[C]// 2019 IEEE/CVF Conference on Computer Vision and Pattern Recognition (CVPR). arXiv, 2019.
\bibitem{b17} Zhang S, Chen C, Hu X, et al. Balanced Knowledge Distillation for Long-tailed Learning[J]. arXiv preprint arXiv:2104.10510, 2021.
\bibitem{b18} Tran D, Wang H, Torresani L, et al. Video classification with channel-separated convolutional networks[C]//Proceedings of the IEEE/CVF International Conference on Computer Vision. 2019: 5552-5561.
\bibitem{b19} Wang H, Tran D, Torresani L, et al. Video modeling with correlation networks[C]//Proceedings of the IEEE/CVF Conference on Computer Vision and Pattern Recognition. 2020: 352-361.
\bibitem{b20} Carreira J, Zisserman A. Quo vadis, action recognition? a new model and the kinetics dataset[C]//proceedings of the IEEE Conference on Computer Vision and Pattern Recognition. 2017: 6299-6308.
\bibitem{b21} Feichtenhofer C, Pinz A, Zisserman A. Convolutional two-stream network fusion for video action recognition[C]//Proceedings of the IEEE conference on computer vision and pattern recognition. 2016: 1933-1941.
\bibitem{b22} Howard A G, Zhu M, Chen B, et al. Mobilenets: Efficient convolutional neural networks for mobile vision applications[J]. arXiv preprint arXiv:1704.04861, 2017.
\bibitem{b23} Tan M, Chen B, Pang R, et al. Mnasnet: Platform-aware neural architecture search for mobile[C]//Proceedings of the IEEE/CVF Conference on Computer Vision and Pattern Recognition. 2019: 2820-2828.
\bibitem{b24} Hu J, Shen L, Sun G. Squeeze-and-excitation networks[C]//Proceedings of the IEEE conference on computer vision and pattern recognition. 2018: 7132-7141.
\bibitem{b25} Woo S, Park J, Lee J Y, et al. Cbam: Convolutional block attention module[C]//Proceedings of the European conference on computer vision (ECCV). 2018: 3-19.
\bibitem{FFmpeg} Ken, Tsutsuguchi. FFmpeg[J]. Journal of the Institute of Image Information \& Television Engineers, 2010.
\bibitem{R-CNN} Ren S ,  He K ,  Girshick R , et al. Faster R-CNN: Towards Real-Time Object Detection with Region Proposal Networks[J]. IEEE Transactions on Pattern Analysis \& Machine Intelligence, 2017, 39(6):1137-1149.
\bibitem{VIA} Dutta A, Zisserman A. The VIA annotation software for images, audio and video[C]//Proceedings of the 27th ACM international conference on multimedia. 2019: 2276-2279.


\end{thebibliography}
\end{document}